\documentclass{article}

\usepackage{times}
\usepackage{graphicx} 
\usepackage{subfigure}

\usepackage{natbib}

\usepackage{algorithm}
\usepackage{algorithmic}

\usepackage{hyperref}

\usepackage[accepted]{icml2014}

\usepackage{amsmath}
\usepackage{cases}
\usepackage{breakurl}

\icmltitlerunning{Can Image-Level Labels Replace Pixel-Level Labels for Image Parsing}

\begin{document}

\twocolumn[
\icmltitle{Can Image-Level Labels Replace Pixel-Level Labels for Image Parsing}

\icmlauthor{Zhiwu Lu}{zhiwu.lu@gmail.com}
\icmladdress{School of Information, Renmin University of China, Beijing 100872, China}
\icmlauthor{Zhenyong Fu and Tao Xiang}{\{z.fu,t.xiang\}@qmul.ac.uk}
\icmladdress{School of EECS, Queen Mary University of London, London, UK}
\icmlauthor{Liwei Wang}{wanglw@cis.pku.edu.cn}
\icmladdress{Key Laboratory of Machine Perception (MOE), School of EECS, Peking University, Beijing 100871, China}
\icmlauthor{Ji-Rong Wen}{jirong.wen@gmail.com}
\icmladdress{School of Information, Renmin University of China, Beijing 100872, China}

\icmlkeywords{Weakly Supervised Learning, Sparse Learning, Noisily Tagged Images, Image Parsing}

\vskip 0.3in
]

\begin{abstract}
This paper presents a weakly supervised sparse learning approach to the problem of noisily tagged image parsing, or segmenting all the objects within a noisily tagged image and identifying their categories (i.e. tags). Different from the traditional image parsing that takes pixel-level labels as strong supervisory information, our noisily tagged image parsing is provided with noisy tags of all the images (i.e. image-level labels), which is a natural setting for social image collections (e.g. Flickr). By oversegmenting all the images into regions, we formulate noisily tagged image parsing as a weakly supervised sparse learning problem over all the regions, where the initial labels of each region are inferred from image-level labels. Furthermore, we develop an efficient algorithm to solve such weakly supervised sparse learning problem. The experimental results on two benchmark datasets show the effectiveness of our approach. More notably, the reported surprising results shed some light on answering the question: can image-level labels replace pixel-level labels (hard to access) as supervisory information for image parsing.
\end{abstract}

\section{Introduction}

Image parsing is an interesting problem in computer vision. The goal of image parsing is to segment all the objects within an image and then identifying their categories (i.e. inferring pixel-level labels). In the past years, image parsing has been drawn much attention \cite{SWR06,YMF07,SJC08,KLT09,LRK09,LRK10,CP11,LLS12,TL13}. Although these methods have been reported to achieve promising results, most
of them take pixels-level labels as the inputs of image parsing. In real-world applications, this supervisory information is time-consuming to access, and fully supervised methods cannot be widely applied in practice.

Recently, many efforts have been made to exploit image-level labels for image parsing \cite{VT07,VB10,VFB11,VFB12,LLL13,ZZZ13}, considering that image-level labels are much easier to access than pixels-level labels. The goal of image parsing is then to infer pixels-level labels from this weak supervisory information. As compared to the traditional image parsing that takes pixels-level labels as inputs, such weakly supervised image parsing is more flexible in real-world applications. However, it is still time-consuming to collect the image-level labels. We hope that such supervisory information can be provided for image parsing automatically.

\begin{figure*}[t]
\vspace{0.04in}
\begin{center}
\includegraphics[width=0.95\textwidth]{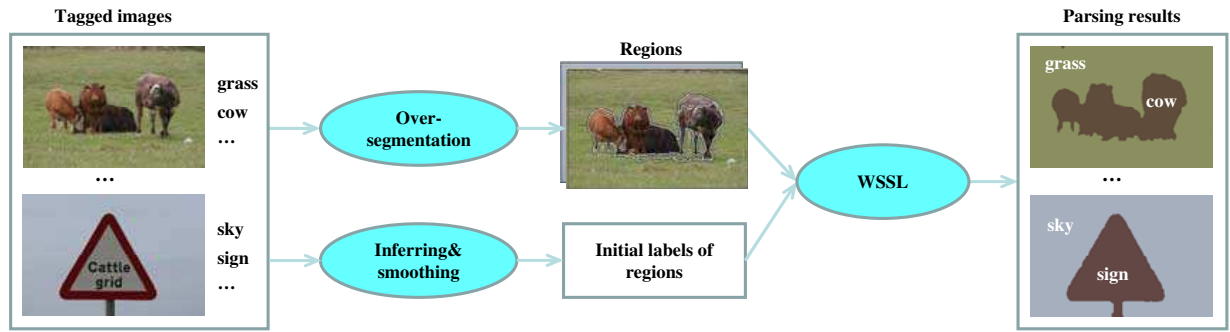}
\end{center}
\vspace{-0.12in} \caption{The flowchart of our weakly supervised sparse learning (WSSL) for image parsing, where the initial labels of regions are inferred from image-level labels and then smoothed by using some prior knowledge about regions and object categories.} \label{Fig.1}
\vspace{-0.05in}
\end{figure*}

Fortunately, this weakly supervised setting for image parsing become feasible with the burgeoning growth of social images over photo-sharing websites (e.g. Flcikr). That is, the tags of social images provided by users can be used as image-level labels for image parsing. It is worth noting that the tags of social images may be noisy (or incomplete) in practice \cite{TYH09}, although they are very easy to access. Hence, the main challenge lies in how to effectively exploit the noisy image-level labels for image parsing. However, such noisily tagged image parsing has been completely ignored in recent work \cite{VT07,VB10,VFB11,VFB12,LLL13,ZZZ13}.

In this paper, we thus focus on developing a weakly supervised sparse learning approach to the problem of noisily tagged image parsing. The basic idea is to first oversegment all the images into regions and then infer the labels of regions from the initial image-level labels. It should be noted that the initial labels of regions cannot be accurately estimated, even if clean image-level labels are initially provided for image parsing. Hence, our main motivation is to continuously suppress the noise in the labels of regions by an iteration procedure. More concretely, inspired by the success use of sparse learning \cite{EA06,MES08,WYG09} for noise reduction in different challenging tasks, we formulate such iteration procedure as a weakly supervised sparse learning problem over all the regions. Here, we do not impose extra requirements on the initial image-level labels, unlike \cite{LLL13} that is limited to clean and complete image-level labels for image parsing. Based on $L_1$-minimization techniques \cite{Donoho04,FNW07,LBR07,CLK10}, we further develop an efficient iterative algorithm to solve such weakly supervised sparse learning problem for noisily tagged image parsing.

The flowchart of our weakly supervised sparse learning (WSSL) for image parsing is illustrated in Figure~\ref{Fig.1}. Here, we adopt the Blobworld method \cite{CBGM02} for oversegmentation, since it can automatically detect the number of regions within an image. Moreover, we utilize the image-level labels provided for image parsing to infer the initial labels of regions, which are further smoothed by using some prior knowledge about regions and object categories. In this paper, to make comparison with the state-of-the-art image parsing results \cite{LRK09,CP11,LLS12,ZZZ13}, we conduct experiments on two benchmark datasets: MSRC \cite{SWR06} and VOC2007 \cite{EVW07}. Meanwhile, to verify the effectiveness of our WSSL approach to noisily tagged image parsing, we then add random noise to the initial image-level labels for the two benchmark datasets. As shown in our later experiments, our WSSL approach can achieve better (or at least comparable) results with respect to the state-of-the-arts. Especially on the VOC2007 dataset, our WSSL approach is shown to have superior performance in image parsing even if we add much noise to the initial image-level labels.

To emphasize our main contributions, we summarize the following distinct advantages of our WSSL approach:
\begin{itemize}
\item This is the first attempt to formulate image parsing as weakly supervised sparse learning from the viewpoint of noise reduction over the labels of regions, to the best of our knowledge. In fact, the challenging problem of noisily tagged image parsing has been ignored in the context of image parsing.
\item We have developed an efficient algorithm for our weakly supervised sparse learning based on $L_1$-minimization techniques, unlike many previous image parsing approaches that incur too large time cost. Moreover, although only tested in image parsing, our WSSL approach can be readily applied to other challenging tasks for the need of noise reduction.
\item The surprising results achieved by our WSSL approach shed some light on answering the question: can image-level labels replace pixel-level labels as supervisory information for image parsing. That is, when the image-level labels (maybe noisy) are easy to access in practice, we can achieve promising image parsing results, without the need to directly collect the pixel-level labels with too expensive cost.
\end{itemize}

The remainder of this paper is organized as follows.
Section~\ref{sect:rw} provides a brief review of related work. In
Section~\ref{sect:wssl}, we develop an efficient weakly supervised sparse learning algorithm based on $L_1$-minimization techniques. In Section~\ref{sect:app}, the proposed algorithm is applied to noisily tagged image parsing. In Sections~\ref{sect:exp} and \ref{sect:con}, we present the experimental results and draw our conclusions, respectively.

\section{Related Work}
\label{sect:rw}

In this paper, we investigate a new type of weakly supervised image parsing which takes the noisy tags of the images (i.e. image-level labels) as the inputs. It is worth noting that our weakly supervised setting is quite different from the other two settings for image parsing that have been widely adopted in the literature. In \cite{SWR06,YMF07,SJC08,KLT09,LRK09,LRK10,CP11,LLS12,TL13}, a fully supervised stetting is considered for image parsing which can achieve promising results by utilizing pixels-level labels for training, but the pixels-level labels are time-consuming to obtain in practice. In \cite{VT07,VB10,VFB11,VFB12,LLL13,ZZZ13}, the image-level labels easier to access are thus utilized for training in image parsing, but some extra requirements are usually imposed on the initial image-level labels. For example, \cite{LLL13} is limited to clean and complete image-level labels due to the special model used for image parsing. In contrast, our noisily tagged image parsing takes the noisy tags of all the images as the supervisory information. This weakly supervised setting is natural for social image collections (e.g. Flickr), where the noisy tags provided by social users are easy to access. Under this setting, the main challenge in image parsing thus lies in how to exploit the noisy image-level labels effectively.

Inspired by the success use of sparse learning for noise reduction in different challenging tasks \cite{EA06,MES08,WYG09}, we thus formulate noisily tagged image parsing as a weakly supervised sparse learning problem. To the best of our knowledge, we have made the first attempt to formulate image parsing from the viewpoint of noise reduction over the labels of regions. In such problem formulation, we actually impose no extra requirements on the initial image-level labels (which can be noisy or even incomplete), unlike the recent work \cite{LLL13}. This means that our weakly supervised sparse learning is more flexible in real-world applications. Moreover, although sparse reconstruction has been applied to image parsing in \cite{ZZZ13}, the problem of noisily tagged image parsing is completely ignored. Finally, it should be noted that sparse learning has been used for noise reduction over the noisy tags of images in the social image classification and retrieval tasks \cite{TYH09}. However, our noisily tagged image parsing is much more challenging than these tasks considering that the initial labels of regions cannot be accurately estimated even if clean image-level labels are provided initially.

Based on $L_1$-minimization techniques \cite{Donoho04,FNW07,LBR07,CLK10}, we develop an efficient algorithm for our weakly supervised sparse learning, unlike many previous image parsing approaches \cite{SWR06,LRK09,VFB11,VFB12} that incur too large time cost during training a generative or discriminative model. When evaluated on two benchmark datasets, our weakly supervised sparse learning algorithm for image parsing can achieve better (or at least comparable) results than the state-of-the-art methods \cite{LRK09,CP11,LLS12,ZZZ13}. Especially on the VOC2007 dataset, our algorithm is shown to outperform the representative fully-supervised image parsing method \cite{LRK09}, even if we add much noise to the initial image-level labels. This means that we actually can exploit the image-level labels (easy to access but maybe noisy) from social image collections for image parsing, instead of the pixel-level labels hard to access in practice.

\section{Weakly Supervised Sparse Learning}
\label{sect:wssl}

In this section, we first formulate noisily tagged image parsing as a weakly supervised learning problem from the viewpoint of noise reduction over the labels of regions. Furthermore, we discuss how to smooth the initial labels of regions which cannot be accurately estimated initially. Finally, we develop an efficient algorithm to solve the weakly supervised sparse learning problem.

\subsection{Problem Formulation}
\label{sect:wssl:prob}

The problem of noisily tagged image parsing is defined as follows. Given a set of noisily tagged images, we adopt the Blobworld method \cite{CBGM02} to oversegment each image and then generate a large set of regions $\mathcal{X}=\{x_1,...,x_N\}$, where $N$ is the total number of regions collected from all the images. In this paper, each region $x_i~(i=1,...,N)$ is denoted by a feature vector by extracting color and texture information from this region. Moreover, the initial labels of all the regions $Y=\{y_{ij}\}_{N\times C}$ are inferred from the image-level labels (i.e. noisy tags) provided for image parsing as: $y_{ij}=1$ if the region $x_i$ belongs to an image which is labeled with category $j$ and $y_{ij}=0$ otherwise, where $C$ is the number of object categories. However, the initial labels of regions cannot be accurately estimated by such simple inference, even if clean image-level labels are provided initially. Hence, we need to perform noise reduction over $Y$ for image parsing. In the following, we will formulate such noise reduction as a weakly supervised sparse learning problem.

Before giving our problem formulation, we first model the whole set of regions as a graph $\mathcal{G}=\{\mathcal{V},W\}$ with its vertex set $\mathcal{V} = \mathcal{X}$ and weight matrix $W=\{w_{ij}\}_{N\times N}$, where $w_{ij}$ denotes the similarity between $x_i$ and $x_j$. The weight matrix $W$ is usually defined by the Gaussian kernel. The normalized Laplacian matrix $\mathcal{L}$ of $\mathcal{G}$ is given by
\begin{eqnarray}
\mathcal{L}=I-D^{-\frac{1}{2}}WD^{-\frac{1}{2}}, \label{eq:lap0}
\end{eqnarray}
where $I$ is an $N\times N$ identity matrix, and $D$ is an $N\times
N$ diagonal matrix with its $i$-th diagonal element being equal to
the sum of the $i$-th row of $W$ (i.e. $\sum_j w_{ij}$). We further derive a new matrix $B$ from
$\mathcal{L}$:
\begin{eqnarray}
B=\Sigma^{\frac{1}{2}}V^T,
\end{eqnarray}
where $V$ is an $N\times N$ orthonormal matrix with each column
being an eigenvector of $\mathcal{L}$, and $\Sigma$ is an $N\times
N$ diagonal matrix with its diagonal element $\Sigma_{ii}$ being an
eigenvalue of $\mathcal{L}$ (sorted as $0\leq \Sigma_{11} \leq
...\leq \Sigma_{NN}$). Since the eigenvalue decomposition of
$\mathcal{L}$ is denoted as $\mathcal{L}=V\Sigma V^T$, we can
represent $\mathcal{L}$ in a symmetrical decomposition form:
\begin{eqnarray}
\mathcal{L}= (\Sigma^{\frac{1}{2}} V^T)^T\Sigma^{\frac{1}{2}} V^T =
B^TB. \label{eq:lap}
\end{eqnarray}

Based on the aforementioned notations, we formulate noisily tagged image parsing as the following weakly supervised sparse learning problem from the viewpoint of noise reduction over the labels of regions:
\begin{eqnarray}
\min_{\hat{Y}\geq 0,F} \frac{1}{2}||\hat{Y}-F||_F^2 + \lambda ||BF||_1 + \gamma||\hat{Y}-Y||_1, \label{eq:ssrc}
\end{eqnarray}
where $\hat{Y} \in R^{N\times C}$ collects the optimal labels of regions, $F \in R^{N\times C}$ denotes an intermediate representation for noise reduction, and $\lambda$ (or $\gamma$) denote the positive regularization parameter. The first and third terms of the above objective function are the $L_2$-norm and $L_1$-norm fitting
constraints, respectively. The second term is the graph smoothness
constraint, which means that the intermediate representation $F$ should not change too much between similar regions. By taking Equation (\ref{eq:lap}) into account, we find that this $L_1$-norm graph smoothness is closely related to the well-known Laplacian  regularization \cite{ZBLW04,ZGL03} which has been widely used for graph-based learning in the literature.

If our weakly supervised sparse learning is considered as compressed sensing, the graph smoothness constraint $||BF||_1 $ actually induces kind of sparsity in the compressed domain spanned by the eigenvectors of $\mathcal{L}$ (see more discussion in Subsection \ref{sect:wssl:alg}). The effect of such sparsity penalty can be transferred to $\hat{Y}$ by solving Equation (\ref{eq:ssrc}) with $F$ being an intermediate representation. Meanwhile, the $L_1$-norm fitting constraint $||\hat{Y}-Y||_1$ imposes direct noise reduction on $Y$. That is, we have induced not only the smoothness sparsity
$||BF||_1$ in the compressed domain but also the noise sparsity
$||\hat{Y}-Y||_1$ in the original space into our weakly supervised sparse learning.

Although we have successfully given the problem formulation for noisily tagged image parsing from the viewpoint of noise reduction, there remain two key problems to be concerned: \emph{how to smooth the initial labels of regions stored in $Y$}, and \emph{how to solve the weakly supervised sparse learning problem efficiently}. In the next two subsections, we will address these two problems, respectively.

\subsection{Initial Label Smoothing}
\label{sect:wssl:smooth}

It should be noted that there exists much noise in the initial labels of regions stored in $Y$ which are estimated only by a simple inference from the image-level labels, even if we are initially provided with clean image-level labels. Such noise issue becomes more serious when the problem of noisily tagged image parsing is considered. In the following, we adopt two smoothing techniques to suppress the noise in $Y$ and reestimate $Y$ as accurately as possible.

We first smooth the initial labels of regions stored in $Y$ by considering the relative size of regions. Let $\rho_i$ be the ratio of the region $x_i$ occupying the image that $x_i$ belongs to, where $i=1,...,N$. We directly define the smoothed labels of regions $\tilde{Y}=\{\tilde{y}_{ij}\}_{N\times C}$ as follows:
\begin{eqnarray}
\tilde{y}_{ij} = \rho_i y_{ij},
\end{eqnarray}
which means that a larger region is more important for our weakly supervised sparse learning. More notably, such smoothing technique can suppress the negative effect of the tiny regions produced by oversegmentation.

We further refine the smoothed $\tilde{Y}$ by exploiting the semantic context of object categories. In the task of image parsing, some object categories may be semantically correlated, e.g., ``water" and ``sky" are at a high probability to occur in the same image while ``water" and "book" are less possible to occur together. In fact, such semantic context can be defined using a single matrix $A=\{a_{jj'}\}_{C\times C}$ based on the Pearson product moment correlation as follows. Let the labels of $M$ images are collected as $Z=\{z_{ij}\}_{M \times C}$, where $z_{ij}=1$ if image $i$ is labeled with category $j$ and $z_{ij}=0$ otherwise. We define $A=\{a_{jj'}\}_{C\times C}$ by:
\begin{eqnarray}
a_{jj'}=\frac{\sum_{i=1}^M
(z_{ij}-\mu_j)(z_{ij'}-\mu_{j'})}{(M-1)\sigma_j\sigma_{j'}},
\end{eqnarray}
where $\mu_j$ and $\sigma_j$ are the mean and standard
deviation of column $j$ of $Z$, respectively. By directly using the label propagation technique \cite{ZBLW04}, we further smooth $\tilde{Y}$ with the semantic context:
\begin{eqnarray}
\bar{Y}=\tilde{Y}(I-\alpha D_c^{-1/2}AD_c^{-1/2})^{-1},
\end{eqnarray}
where $\alpha\in(0,1)$, and $D_c$ is a diagonal matrix with its $j$-th diagonal element being $\sum_{j'} a_{jj'}$.

Once we have obtained the final smoothed $\bar{Y}$, we reformulate our weakly supervised sparse learning problem as:
\begin{eqnarray}
\min_{\hat{Y}\geq 0,F} \frac{1}{2}||\hat{Y}-F||_F^2 + \lambda ||BF||_1 + \gamma||\hat{Y}-\bar{Y}||_1. \label{eq:wssl}
\end{eqnarray}
In the following, we will develop an efficient algorithm to solve the above problem for noisily tagged image parsing.

\subsection{Efficient WSSL Algorithm}
\label{sect:wssl:alg}

The $L_1$-minimization problem (\ref{eq:wssl}) can be solved in the following two alternate optimization steps:
\begin{eqnarray}
\hspace{-0.25in}&&F^*=\arg\min_F \frac{1}{2}||F-\hat{Y}^*||_F^2 + \lambda ||BF||_1, \label{eq:wssl1} \\
\hspace{-0.25in} && \hat{Y}^* =\arg\min_{\hat{Y}\geq 0} \frac{1}{2}||\hat{Y}-F^*||_F^2 + \gamma||\hat{Y}-\bar{Y}||_1. \label{eq:wssl2}
\end{eqnarray}
We set $\hat{Y}^*=\bar{Y}$ initially. As a basic $L_1$-minimization problem, the second subproblem (\ref{eq:wssl2}) has an explicit solution:
\begin{eqnarray}
\hat{Y}^*=\mathrm{soft\_thr}(F^*,\bar{Y},\gamma), \label{eq:soft}
\end{eqnarray}
where $\mathrm{soft\_thr}(\cdot,\cdot,\gamma)$ is a soft-thresholding
function. Here, we directly define $z=\mathrm{soft\_thr}(x, y, \lambda)$ as:
\begin{eqnarray}
z =\begin{cases} z_1=\max(x-\lambda,y), & f_1\leq f_2 \\
z_2=\max(0,\min(x+\lambda,y)), & f_1> f_2 \\
\end{cases},
\end{eqnarray}
where $f_1=(z_1-x)^2 + 2\lambda|z_1-y|$ and $f_2=(z_2-x)^2 + 2\lambda|z_2-y|$. In the following, we focus on developing an
efficient algorithm to solve the first subproblem (\ref{eq:wssl1}).

It should be noted that directly solving the $L_1$-minimization subproblem (\ref{eq:wssl1}) is computationally intractable in the task of image parsing consider that only the computation of $B$ would incur too large time cost. Fortunately, the dimension of our weakly supervised sparse learning can be reduced dramatically by working only with a small subset of eigenvectors of $\mathcal{L}$. That is, similar to \cite{FWT10}, we significantly reduce the dimension of $F$ by requiring it to take the form of $F = V_{m}U$ where $U$ is an $m \times C$ matrix that collects the reconstruction coefficients and $V_{m}$ is an $N \times m$ matrix whose columns are the $m$ eigenvectors with smallest eigenvalues (i.e. the first $m$ columns of $V$). The first subproblem (\ref{eq:wssl1}) can now be formulated as:
\setlength{\arraycolsep}{0.1em}
\begin{eqnarray}
\hspace{-0.25in}&&\arg\min_U \frac{1}{2}||V_{m}U-\hat{Y}^*||_F^2 + \lambda ||BV_{m}U||_1 \nonumber\\
\hspace{-0.25in}&=&\arg\min_U \sum_{j=1}^C \frac{1}{2}||V_{m}U_{.j} -
\hat{Y}^*_{.j}||_2^2+ \lambda ||BV_{m}U_{.j}||_1, \label{eq:allsc}
\end{eqnarray}
where $\hat{Y}^*_{.j}$ and $U_{.j}$ denote the $j$-th column of
$\hat{Y}^*$ and $U$, respectively. The above problem can be further
decomposed into the following $C$ independent subproblems:
\begin{eqnarray}
&&\arg\min_{U_{.j}} \frac{1}{2}||V_{m}U_{.j}-\hat{Y}^*_{.j}||_2^2+
\lambda ||BV_{m}U_{.j}||_1 \nonumber\\
&=&\arg\min_{U_{.j}} \frac{1}{2}||V_{m}U_{.j}-\hat{Y}^*_{.j}||_2^2+
\lambda ||\sum_{i=1}^m \Sigma^{\frac{1}{2}}V^T V_{.i}u_{ij}||_1 \nonumber\\
&=&\arg\min_{U_{.j}} \frac{1}{2}||V_{m}U_{.j}-\hat{Y}^*_{.j}||_2^2+
\lambda \sum_{i=1}^m \Sigma^{\frac{1}{2}}_{ii} |u_{ij}|,
\label{eq:sc}
\end{eqnarray}
where the orthonormality of $V$ is used to simplify the $L_1$-norm sparsity regularization term $||BV_{m}U_{.j}||_1$. The first term of the above objective function denotes the linear reconstruction error which just takes the same form as that used in the well-known sparse coding \cite{LBR07}, while the second term denotes the weighted $L_1$-norm sparsity regularization over the reconstruction coefficients. That is, the first subproblem (\ref{eq:wssl1}) has been transformed into a generalized sparse coding problem. In this paper, we solve this generalized sparse coding problem directly using the Matlab toolbox L1General\footnote{\url{http://www.di.ens.fr/~mschmidt/Software/L1General.html}}.

The formulation $F_{.j} = V_{m}U_{.j}~(m\ll N)$ used in Equation
(\ref{eq:sc}) has two distinct advantages. Firstly, we can transform the original $L_1$-minimization problem (\ref{eq:wssl1}) into a generalized sparse coding problem, which can then be solved at a linear time cost with respect to the data size $N$. Secondly, we do not need to compute the full matrix $B$, which is especially suitable for image parsing with a large set of regions. In fact, we only need to compute the $m$ smallest eigenvectors of $\mathcal{L}$. To speed up this step, we choose to construct $k$-nearest neighbors ($k$-NN) graphs for our weakly supervised sparse learning. Given a $k$-NN graph ($k\ll N$), the time complexity of finding $m$ smallest eigenvectors of the sparse matrix $\mathcal{L}$ is $O(m^3+m^2N+kmN)$, which is scalable with respect to the data size $N$.

Considering the two alternate optimization steps together, our efficient algorithm for weakly supervised sparse learning (WSSL) is outlined as follows:
\begin{description}
\item[(1)]
Construct a $k$-NN graph with its weight matrix $W$ being defined
over all the regions;
\item[(2)]
Compute the normalized Laplacian matrix $\mathcal{L}=
I-D^{-\frac{1}{2}}WD^{-\frac{1}{2}}$ according to Equation
(\ref{eq:lap0});
\item[(3)]
Find the $m$ smallest eigenvectors of the normalized Laplacian
matrix $\mathcal{L}$ and store them in $V_m$;
\item[(4)]
Collect the initial image-level labels provided for image parsing into $Y$ and then smooth it as $\bar{Y}$;
\item[(5)]
Initialize $\hat{Y}^*$ with the smoothed $\bar{Y}$ as $\hat{Y}^*=\bar{Y}$;
\item[(6)]
Find the best solution $U^*$ of the $L_1$-minimization problem (\ref{eq:allsc}) by running the L1General software with respect to each object category;
\item[(7)]
Update $\hat{Y}^*$ with $F^* = V_mU^*$ according to Equation (\ref{eq:soft}): $\hat{Y}^*= \mathrm{soft\_thr} (F^*,\bar{Y},\gamma)$ ;
\item[(8)]
Iterate Steps 6 and 7 until the stopping condition is satisfied, and output the final parsing results $\hat{Y}^*$.
\end{description}
In our later experiments, we find that our WSSL algorithm generally
converges in very limited number of iteration steps ($<5$).
Moreover, according to our aforementioned analysis, both Steps 3 and
6 are scalable with respect to the data size $N$. Hence, our WSSL
algorithm can be applied to image parsing with a large set of regions.

\subsection{Discussion}
\label{sect:wssl:dis}

Although sparse learning has been successfully used for noise reduction in different tasks \cite{EA06,MES08,WYG09}, little attention has been paid to noisily tagged image parsing based on sparse learning. In this paper, we develop a weakly supervised sparse learning algorithm for noisily tagged image parsing. In fact, we have made the first attempt to formulate image parsing as weakly supervised sparse learning from the viewpoint of noise reduction over the labels of regions. In our problem formulation, we impose no extra requirements on the initial image-level labels (which can be noisy or even incomplete) provided for image parsing, unlike the recent work \cite{LLL13}. This means that our weakly supervised sparse learning is more flexible in real-world applications. More notably, although originally designed for noisily tagged image parsing, our weakly supervised sparse learning can be readily applied to other challenging tasks for the need of noise reduction.

\section{Noisily Tagged Image Parsing}
\label{sect:app}

In the last section, we have just developed an efficient weakly supervised sparse learning algorithm for noisily tagged image parsing. The inputs of this algorithm are a large set of regions and a set of image-level labels. To pay our main attention to algorithm design itself, we assume that the large set of regions have been provided in advance. In the following, we will discuss how to generate this input for our weakly supervised sparse learning algorithm.

Given a set of images, we adopt the Blobworld method \cite{CBGM02} for oversegmentation. Concretely, we first extract a 6-dimensional vector of color and texture features for each pixel of an image and then model this image as a Gaussian mixture model. The pixels within this image are then grouped into regions, and the number of regions is automatically detected by a model selection principle. To ensure an oversegmentation of each image, we slightly modify the original Blobworld method: 1) the number of regions is initially set to a large value; 2) model selection is forced to be less important during segmentation.

After we have oversegmented all the images, we generate a large set of regions $\mathcal{X}=\{x_1,...,x_N\}$, where each region is denoted as a 137-dimensional feature vector by concatenating color and textual visual features. The 137-dimensional feature vector includes: three mean color features with their standard deviations (6-dimensional), three mean texture features with their standard deviations (6-dimensional), and 125-dimensional color histogram. Finally, we compute a Gaussian kernel over $\mathcal{X}$ for our weakly supervised sparse learning algorithm.

The full algorithm for noisily tagged image parsing can be
summarized as follows:
\begin{description}
\item[(1)]
Oversegment each image into multiple regions using the modified Blobworld method;
\item[(2)]
Extract a 137-dimensional feature vector from each region by concatenating color and textual features;
\item[(3)]
Perform image parsing by running our weakly supervised sparse learning algorithm proposed in Subsection~\ref{sect:wssl:alg} over the large set of regions.
\end{description}
As we have mentioned in Subsection~\ref{sect:wssl:alg}, Step (3) for image parsing runs very efficiently even on a large set of regions, unlike many previous image parsing approaches \cite{SWR06,LRK09,VFB11,VFB12} that incur too large time cost during training a generative or discriminative model. Moreover, although Step (1) is the most time-consuming in the above algorithm, we can readily run this step in a distributed computing way to speed it up.

\section{Experimental Results}
\label{sect:exp}

In this section, we evaluate our weakly supervised sparse learning (WSSL) algorithm for noisily tagged image parsing. We first describe the experimental setup and then compare our algorithm with other closely related methods.

\subsection{Experimental Setup}

\begin{figure*}[t]
\vspace{0.04in}
\begin{center}
\includegraphics[width=0.98\textwidth]{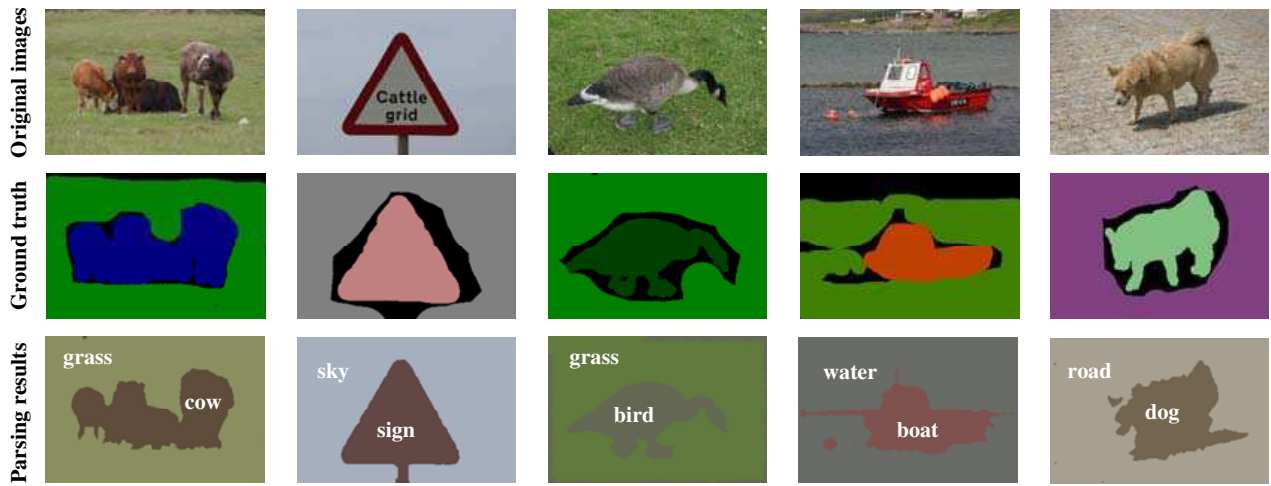}
\end{center}
\vspace{-0.15in}
\caption{Some example results for image parsing by our WSSL algorithm in comparison with the ground-truth on the MSRC dataset.}
\label{Fig.2}
\vspace{-0.15in}
\end{figure*}

We select two benchmark datasets for performance evaluation: MSRC \cite{SWR06}, and VOC2007 \cite{EVW07}. The MSRC benchmark dataset contains 591 images, accompanied with a hand labeled object segmentation of 21 object categories. Pixels on the boundaries of objects are usually labeled as background and not taken into account in these segmentations. The PASCAL VOC2007 benchmark dataset was used for the PASCAL Visual Object Category segmentation contest 2007. It contains 5,011 training and 4,952 testing images where only the bounding boxes of the objects present in the image are marked, and 20 object classes are given for the task of classification, detection, and segmentation. Rather than on the 5,011 annotated training images with bounding box indicating object location and rough boundary, we conduct experiments on the segmentation set with the `train-val' split including 632 images, which are well segmented and thus are suitable for the evaluation of image parsing. For the two benchmark datasets, we oversegment each image into multiple regions and then totally obtain about 7,000 regions and 15,000 regions, respectively.

\begin{table}[t]
\vspace{-0.04in} \caption{Accuracies (\%) of different learning methods for noisily tagged image parsing on the MSRC dataset. The standard deviations are also provided along with the average accuracies.}
\label{Table.1} \vspace{-0.02in}
\begin{small}
\begin{center}
\tabcolsep0.14cm
\begin{tabular}{|l|ccccc|}
\hline
Noisily tagged images & 0\% & 25\% & 50\% & 75\% & 100\% \\
\hline
Our WSSL        & \textbf{76}$\pm$0 & \textbf{70}$\pm$1 & \textbf{65}$\pm$1 & \textbf{60}$\pm$2 & \textbf{55}$\pm$2\\
\cite{ZBLW04}& 68$\pm$0 & 63$\pm$1 & 58$\pm$1 & 53$\pm$1 & 46$\pm$2\\
\cite{TYH09} & 70$\pm$0 & 63$\pm$1 & 56$\pm$1 & 50$\pm$1 & 43$\pm$2 \\
\cite{CLK10} & 51$\pm$0 & 50$\pm$1 & 50$\pm$1 & 49$\pm$1 & 47$\pm$1\\
\hline
\end{tabular}
\end{center}
\end{small}
\vspace{-0.17in}
\end{table}

To verify the effectiveness of our WSSL algorithm for noisily tagged image parsing, we add random noise to the image-level labels provided initially. Concretely, we randomly select certain percent of images and then attach each selected image with an extra wrong label. That is, the noise strength is determined by the percent of noisily tagged images in the following experiments. By considering image parsing as noise reduction over $\bar{Y}$, we compare our WSSL algorithm with two representative sparse learning methods \cite{TYH09,CLK10} in the literature. Since the sparsity regularization $||BF||_1$ used by our WSSL algorithm is related to the well-known Laplacian regularization, we also make comparison to \cite{ZBLW04} which is a representative method using Laplacian regularization. Moreover, to compare our WSSL algorithm with the state-of-the-art methods, we evaluate the image parsing results on a subset of images equivalent to the test set used in \cite{LRK09,LLS12,SJC08,VFB11,ZZZ13} for each benchmark dataset. In this paper, the accuracies are used to measure the image parsing results by comparing them to the ground truth segmentations pixel by pixel, which are further averaged over 25 random runs.

In the experiments, we construct $k$-NN graphs over the large set of regions $\mathcal{X}$ for all the related methods to speed up image parsing. Moreover, we set the parameters of our WSSL algorithm as $k=550$, $\alpha=0.05$, $\lambda=0.01$, $\gamma=0.01$, and $m=35$ uniformly for the two benchmark datasets. The parameters of other closely related methods are also set their respective optimal values.

\begin{table}[t]
\vspace{-0.04in} \caption{Accuracies (\%) of different learning methods for noisily tagged image parsing on the VOC2007 dataset. The standard deviations are also provided along with the average accuracies.}
\label{Table.2} \vspace{-0.02in}
\begin{small}
\begin{center}
\tabcolsep0.14cm
\begin{tabular}{|l|ccccc|}
\hline
Noisily tagged images & 0\% & 25\% & 50\% & 75\% & 100\% \\
\hline
Our WSSL        & \textbf{49}$\pm$0 & \textbf{45}$\pm$1 & \textbf{40}$\pm$1 & \textbf{36}$\pm$2 & \textbf{33}$\pm$2\\
\cite{ZBLW04}& 28$\pm$0 & 25$\pm$1 & 23$\pm$1 & 20$\pm$2 & 17$\pm$1\\
\cite{TYH09} & 34$\pm$0 & 31$\pm$1 & 27$\pm$1 & 24$\pm$2 & 20$\pm$1 \\
\cite{CLK10} & 31$\pm$0 & 28$\pm$1 & 25$\pm$1 & 21$\pm$1 & 18$\pm$2\\
\hline
\end{tabular}
\end{center}
\end{small}
\vspace{-0.18in}
\end{table}

\subsection{Parsing Results}

\begin{table*}[t]
\vspace{-0.00in} \caption{Accuracies (\%) of different image parsing methods (with three settings) for individual categories on the MSRC benchmark dataset. The last column is the average accuracy over all the categories.}
\label{Table.3} \vspace{-0.02in}
\begin{small}
\begin{center}
\tabcolsep0.111cm
\begin{tabular}{|l|l|ccccccccccccccccccccc|c|}
\hline
Settings & Methods  & \rotatebox{90}{building} & \rotatebox{90}{grass} & \rotatebox{90}{tree} & \rotatebox{90}{cow} & \rotatebox{90}{sheep} & \rotatebox{90}{sky} & \rotatebox{90}{aeroplane} & \rotatebox{90}{water} & \rotatebox{90}{face} & \rotatebox{90}{car} & \rotatebox{90}{bicycle} & \rotatebox{90}{flower} & \rotatebox{90}{sign} & \rotatebox{90}{bird} & \rotatebox{90}{book} & \rotatebox{90}{chair} & \rotatebox{90}{road} & \rotatebox{90}{cat} & \rotatebox{90}{dog} & \rotatebox{90}{body} & \rotatebox{90}{boat}  & \rotatebox{90}{average} \\
\hline
Fully & \cite{LRK09} & \textbf{80} & \textbf{96} & 86 & 74 & 87 & \textbf{99} & 74 & \textbf{87} & 86 & 87 & 82 & 97 & \textbf{95} & 30 & 86 & 31 & \textbf{95} & 51 & 69 & 66 & 9 & 75 \\
supervised & \cite{LLS12} & 59 & 90 & \textbf{92} & 82 & 83 & 94 & 91 & 80 & 85 & 88 & 96 & 89 & 73 & 48 & 96 & 62 & 81 & 87 & 33 & 44 & 30 & \textbf{76} \\ \hline
Weakly  & \cite{VFB11}& 5 & 80 & 58 & 81 & \textbf{97} & 87 & \textbf{99} & 63 & 91 & 86 & \textbf{98} & 82 & 67 & 46 & 59 & 45 & 66 & 64 & 45 & 33 & 54 & 67\\
supervised & \cite{ZZZ13}  &  63 & 93 & \textbf{92} & 62 & 75 & 78 & 79 & 64 & \textbf{95} & 79 & 93 & 62 & 76 & 32 & 95 & 48 & 83 & 63 & 38 & \textbf{68} & 15 & 69 \\ \hline
& Ours (0\% noise) & 17	& 38 & 64 & \textbf{96} & 96 & 56 & 92 & 72 & 87 & \textbf{90} & 77 & \textbf{98} & \textbf{95} & \textbf{94} & \textbf{97} & \textbf{100} & 31 & \textbf{99} & \textbf{94} & 23 & \textbf{75} & \textbf{76} \\
Noisily & Ours (25\% noise) & 17 & 37 & 57 & 87 & 90 & 53& 88 & 68 & 70 & 81 & 73 & 93 & 88 & 87 & 93 & 97 & 28 & 94 & 86 & 16 & 73 & 70 \\
tagged & Ours (50\% noise) & 12 & 35 & 50 & 77 & 81 & 50 & 84 & 65 & 62 & 70 & 74 & 85 & 84 & 77 & 86 & 90 & 25 & 91 & 84 & 14 & 69 & 65 \\
& Ours (75\% noise) & 7 & 34 & 43 & 63 & 73 & 49 & 78 & 60 & 53 & 63 & 75 & 79 & 80 & 69 & 82 & 88 & 24 & 88 & 80 & 9 & 62 & 60 \\
\hline
\end{tabular}
\end{center}
\end{small}
\vspace{-0.05in}
\end{table*}

\begin{table*}[t]
\vspace{-0.00in} \caption{Accuracies (\%) of different image parsing methods (with three settings) for individual categories on the VOC2007 benchmark dataset. The last column is the average accuracy over all the categories.}
\label{Table.4} \vspace{-0.02in}
\begin{small}
\begin{center}
\tabcolsep0.135cm
\begin{tabular}{|l|l|cccccccccccccccccccc|c|}
\hline
Settings & Methods  & \rotatebox{90}{aeroplane} & \rotatebox{90}{bicycle} & \rotatebox{90}{bird} & \rotatebox{90}{boat} & \rotatebox{90}{bottle} & \rotatebox{90}{bus} & \rotatebox{90}{car} & \rotatebox{90}{cat} & \rotatebox{90}{chair} & \rotatebox{90}{cow} & \rotatebox{90}{diningtable} & \rotatebox{90}{dog} & \rotatebox{90}{horse} & \rotatebox{90}{motorbike} & \rotatebox{90}{person} & \rotatebox{90}{pottedplant} & \rotatebox{90}{sheep} & \rotatebox{90}{sofa} & \rotatebox{90}{train} & \rotatebox{90}{tvmonitor} & \rotatebox{90}{average} \\
\hline
Fully & \cite{LRK09} & 27 & 33 & 44 & 11 & 14 & 36 & 30 & 31 & 27 & 6 & \textbf{50} & 28 & 24 & 38 & \textbf{52} & 29 & 28 & 12 & 45 & 46 & 30 \\
supervised & TKK from VOC2007 & 19 & 21 & 5 & 16 & 3 & 1 & \textbf{78} & 1 & 3 & 1 & 23 & \textbf{69} & 44 & 42 & 0 & \textbf{65} & 30 & 35 & \textbf{89} & \textbf{71} & 31 \\ \hline
Weakly  & \cite{SJC08}& 14 & 8 & 11 & 0 & 17 & \textbf{46} & 5 & 13 & 4 & 0 & 30 & 29 & 12 & 18 & 40 & 6 & 17 & 17 & 14 & 9 & 16\\
supervised & \cite{ZZZ13}  &  48 & 20 & 26 & 25 & 3 & 7 & 23 & 13 & 38 & 19 & 15 & 39 & 17 & 18 & 25 & 47 & 9 & \textbf{41} & 17 & 33 & 24 \\ \hline
& Ours (0\% noise) & \textbf{54} & \textbf{69} & \textbf{52} & \textbf{50} & \textbf{64} & \textbf{46} & 41 & \textbf{51} & \textbf{37} & \textbf{57} & 16 & 47 & \textbf{76} & \textbf{77} & 26 & 36 & \textbf{47} & 25 & 83 & 25 & \textbf{49} \\
Noisily  & Ours (25\% noise) & 49 & 66 & 44 & 47 & 62 & 44 & 39 & 44 & 32 & 51 & 16 & 44 & 66 & 72 & 25 & 31 & 42 & 20 & 78 & 22 & 45 \\
tagged& Ours (50\% noise) & 46 & 61 & 36 & 40 & 59 & 38 & 37 & 38 & 32 & 43 & 16 & 40 & 57 & 67 & 21 & 30 & 36 & 20 & 72 & 18 &  40 \\
& Ours (75\% noise) & 40 & 52 & 30 & 39 & 56 & 35 & 33 & 37 & 28 & 33 & 15 & 40 & 49 & 60 & 18 & 26 & 32 & 21 & 63 & 16 & 36 \\
\hline
\end{tabular}
\end{center}
\end{small}
\vspace{-0.05in}
\end{table*}

We first compare our WSSL algorithm to \cite{ZBLW04,TYH09,CLK10} by considering noisily tagged image parsing as noise reduction over $\bar{Y}$. The comparison results on the two benchmark datasets are listed in Tables~\ref{Table.1} and \ref{Table.2}, respectively. Here, different percents of noisily tagged images are provided for noisily tagged image parsing initially. The immediate observation is that our WSSL algorithm achieves the best results in all cases (see some example results shown in Figure~\ref{Fig.2}). This means that the two types of sparsity regularization used by our WSSL algorithm indeed help to suppress the noise in the initial labels of regions for noisily tagged image parsing. Here, it should be noted that only one type of sparsity regularization has been considered in \cite{TYH09,CLK10} for sparse learning. Moreover, the impressive results reported in these two tables also demonstrate that the main contributions of this paper lie in not only our new problem formulation for image parsing but also our effective WSSL algorithm itself.

We further show the comparison of our WSSL algorithm to the state-of-the-art image parsing methods on the two benchmark datasets in Tables~\ref{Table.3} and \ref{Table.4}, respectively. Here, three settings for image parsing are considered: fully supervised image parsing using pixel-level labels \cite{LRK09,LLS12}, weakly supervised image parsing using clean image-level labels \cite{VFB11,ZZZ13,SJC08}, and our noisily tagged image parsing using noisy image-level labels. It can be observed that our WSSL algorithm can achieve better (or at least comparable) results with respect to the state-of-the-arts. In particular, by making observation over individual categories, we find  that our WSSL algorithm performs the best for about half of categories on each benchmark dataset. More notably, for the VOC2007 dataset, our WSSL algorithm is shown to have superior performance even if we add much noise to the initial image-level labels. Such surprising results actually shed some light on answering the question: can image-level labels replace pixel-level labels as supervisory information for image parsing. That is, when the image-level labels (maybe noisy) are easy to access in practice, we can achieve promising image parsing results by our WSSL algorithm, without the need to directly collect the pixel-level labels with too expensive cost.

Besides the above advantages in noisily tagged image parsing, our WSSL algorithm has another distinct advantage, i.e., it runs efficiently even on a large set of regions. For example, the running time of noise reduction over $\bar{Y}$ taken by our WSSL algorithm, \cite{ZBLW04}, \cite{TYH09}, and \cite{CLK10} on the VOC2007 dataset ($N\approx 15,000$) is 40, 47, 145, and 60 seconds, respectively. We run all the algorithms (Matlab code) on a computer with 3GHz CPU and 32GB RAM. It can be clearly observed that our WSSL algorithm runs the fastest among all the four methods for noisily tagged image parsing.

\section{Conclusions}
\label{sect:con}

In this paper, we have investigated the challenging problem of noisily tagged image parsing. From the viewpoint of noise reduction over the labels of regions, we have formulated noisily tagged image parsing as a  weakly supervised sparse learning problem. Based on $L_1$-minimization techniques, we have further developed an efficient algorithm to solve such weakly supervised sparse learning problem. The experimental results on two benchmark datasets have demonstrated the effectiveness of our weakly supervised sparse learning approach to noisily tagged image parsing. The present work can be further improved in two ways: 1) we can first smooth the image-level labels over all the images based on the sophisticated image analysis techniques, before providing them for noisily tagged image parsing as the inputs; 2) we can evaluate our weakly supervised sparse learning approach directly over the social image collections, and we need to generate the ground truth segmentations for the evaluation of image parsing.

\small{

}

\end{document}